%% file: main.tex
\documentclass[10pt,twocolumn,letterpaper]{article}

\usepackage{cvpr}

\definecolor{cvprblue}{rgb}{0.21,0.49,0.74}
\usepackage[pagebackref,breaklinks,colorlinks,allcolors=cvprblue]{hyperref}
\usepackage{ragged2e} 
\usepackage{booktabs,makecell, multirow, tabularx}
\usepackage{pifont}
\usepackage{xcolor}
\usepackage{needspace}
\usepackage{booktabs}
\usepackage{graphicx}
\usepackage{amsmath}
\usepackage{amssymb}
\usepackage{booktabs}
\usepackage{bm}
\usepackage{float}
\usepackage{multirow}
\usepackage{placeins}
\usepackage{colortbl}     
\usepackage[table]{xcolor} 
\definecolor{best}{HTML}{A9DFBF}   
\definecolor{second}{HTML}{D4EFDF}

\usepackage{bbding} 
 
\title{N3D-VLM: Native 3D Grounding Enables Accurate Spatial Reasoning \\ in Vision-Language Models}

\author{
    Yuxin Wang$^{1,2}$\footnotemark[1]\quad
    Lei Ke$^{2}$\quad
    Boqiang Zhang$^{2}$\quad
    Tianyuan Qu$^{2,3}$\quad
    Hanxun Yu$^{2,4}$\quad \\
    Zhenpeng Huang$^{2,5}$\quad
    Meng Yu$^{2}$\quad
    Dan Xu$^{1}$\footnotemark[2]\quad
    Dong Yu$^{2}$ 
    \\ \\
    $^{1}$HKUST \quad $^{2}$Tencent AI Lab \quad $^{3}$CUHK \quad $^{4}$ZJU \quad $^{5}$NJU\\
    \\
    \href{https://n3d-vlm.github.io}{ \color{black} \textbf{Project Page:} https://n3d-vlm.github.io}
}


\begin{document}
\twocolumn[{

\includegraphics[height=3.0em, trim=0cm 0cm 0cm 0cm, clip]{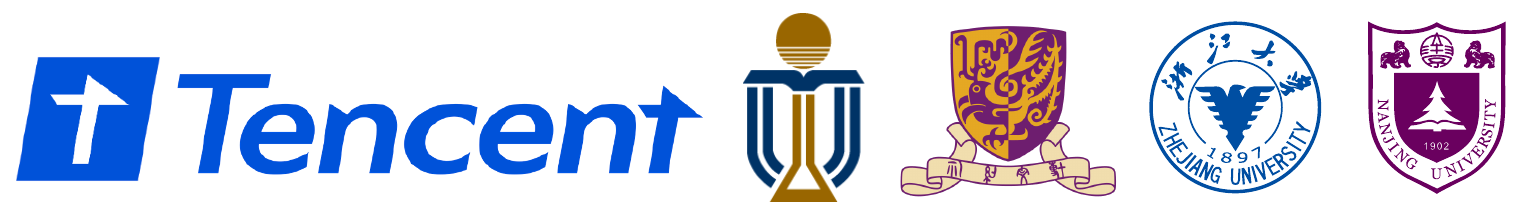}
\hrule height 1pt 
\vspace{4pt} 

\maketitle
  \centering
  \includegraphics[width=1.0\linewidth]{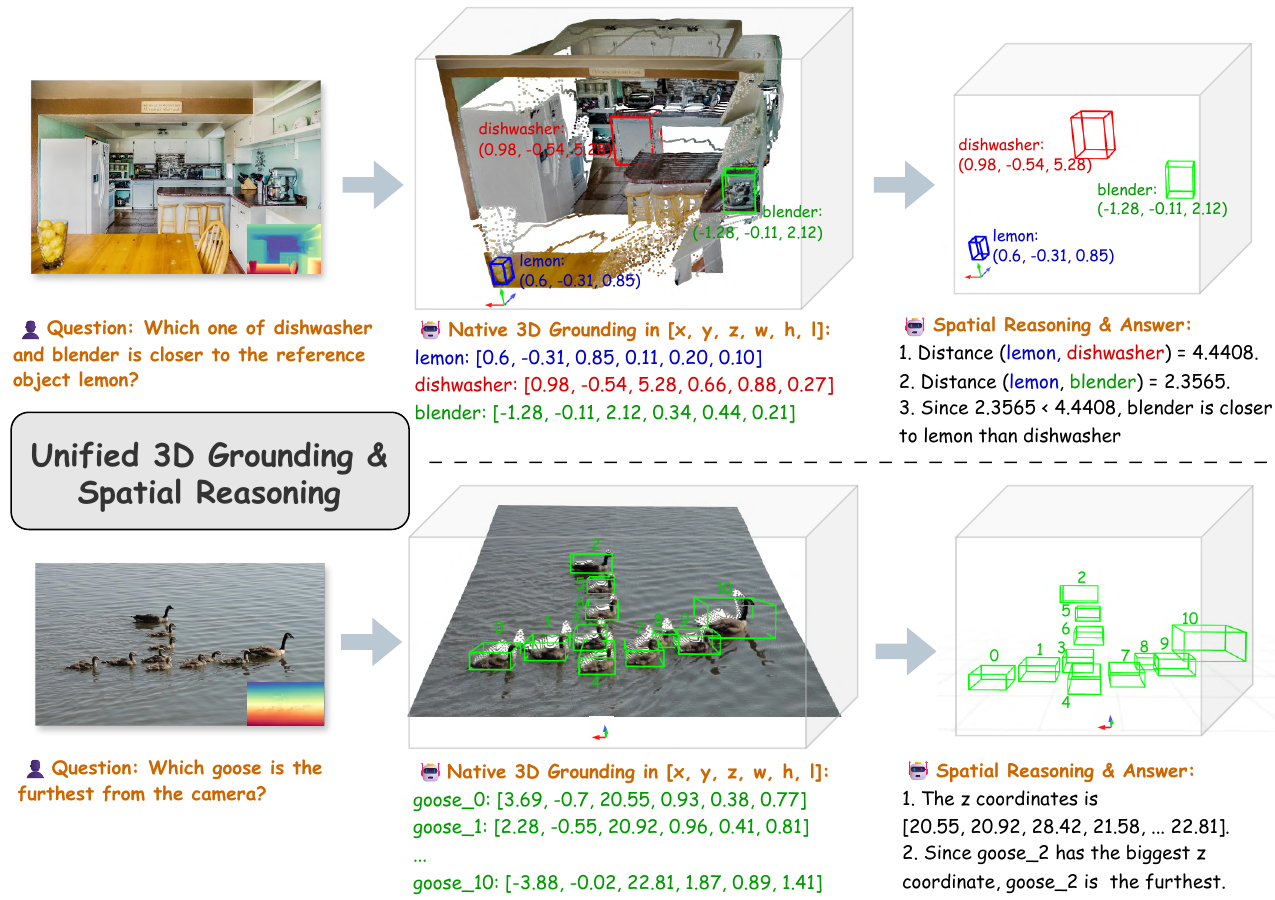}
   \vspace{-17pt}
   \captionsetup{type=figure}
    \captionof{figure}{
    Our unified vision-language model \textbf{N3D-VLM} performs native 3D grounding and subsequent spatial reasoning and answering. Given an RGB image and the corresponding text question, the model is capable of predicting 3D bounding boxes for specified objects and explicitly reasoning about spatial relations in 3D space. 
    }\label{fig:teaser}
   \vspace{5pt}
}]

\renewcommand{\thefootnote}{\fnsymbol{footnote}}
\footnotetext[1]{Work done during an internship at Tencent AI Lab.}
\footnotetext[2]{Corresponding author.}

\input{0_abstract}    
\input{1_intro}
\input{2_related}

\input{3_method}
\input{4_exp}
\input{5_conclusion}
\newpage
{
    \small

\input{main.bbl}
    % \bibliographystyle{ieeenat_fullname}
    % \bibliography{main}
}

\input{X_suppl}

\end{document}

%% file: 0_abstract.tex
\begin{abstract}
While current multimodal models can answer questions based on 2D images, they lack intrinsic 3D object perception, limiting their ability to comprehend spatial relationships and depth cues in 3D scenes.
In this work, we propose \textbf{N3D-VLM}, a novel unified framework that seamlessly integrates native 3D object perception with 3D-aware visual reasoning, enabling both precise 3D grounding and interpretable spatial understanding.
Unlike conventional end-to-end models that directly predict answers from RGB/RGB-D inputs, our approach equips the model with native 3D object perception capabilities, enabling it to directly localize objects in 3D space based on textual descriptions. Building upon accurate 3D object localization, the model further performs explicit reasoning in 3D, achieving more interpretable and structured spatial understanding. 
To support robust training for these capabilities, we develop a scalable data construction pipeline that leverages depth estimation to lift large-scale 2D annotations into 3D space, significantly increasing the diversity and coverage for 3D object grounding data, yielding over six times larger than the largest existing single-image 3D detection dataset. 
Moreover, the pipeline generates spatial question-answering datasets that target chain-of-thought (CoT) reasoning in 3D, facilitating joint training for both 3D object localization and 3D spatial reasoning. 
Experimental results demonstrate that our unified framework not only achieves state-of-the-art performance on 3D grounding tasks, but also consistently surpasses existing methods in 3D spatial reasoning in vision-language model.

\end{abstract}

%% file: 1_intro.tex
\section{Introduction}
\label{sec:intro}

Recent vision-language models (VLMs) \cite{hurst2024gpt,team2024gemini,liu2023llava,bai2025qwen2} have expanded beyond text-only understanding to handle diverse multimodal tasks such as image and video analysis, OCR, and visual reasoning. However, real-world applications often demand a deeper grasp of 3D structure and spatial relationships, which current VLMs largely lack. Effective 3D spatial reasoning requires accurate object-level perception in 3D space; without it, models struggle to infer spatial configurations or reason about physical environments. 
Advancing toward truly multimodal intelligence therefore requires moving beyond 2D language-centric perception toward robust 3D spatial ability to perceive, ground, and reason about the 3D world from visual inputs.

Specialized VLMs enhance 3D spatial understanding capability through diverse input modalities and architectural designs.  
Some models integrate external perception models~\cite{ravi2024sam, schult2022mask3d} to obtain auxiliary object information, such as 2D/3D bounding boxes or segmentation masks~\cite{cheng2024spatialrgpt,xu2024vlmgrounder,he2025spatialormllm}. Others assume that 3D object bounding boxes or spatial layouts are provided in advance~\cite{qi2025gpt4scene, li2025seeground, zheng2025video3dllm}. 
Alternatively, recent approaches have explored using VLMs to directly localize objects in point clouds~\cite{avetisyan2024scenescript,mao2025spatiallm}. However, these methods typically focus on object detection in constrained scenes with limited object categories, and do not support explicit spatial reasoning. 
Although these approaches have advanced specific aspects of 3D spatial understanding, they either depend heavily on external modules or predefined spatial information, or remain confined to narrow perception tasks, which makes it challenging to generalize and integrate them into unified vision-language systems. 

Based on these observations, we argue that 3D spatial understanding could be decomposed into two core abilities: \textbf{3D object localization} and subsequent \textbf{3D spatial reasoning}. 
This perspective motivates our design, where explicit 3D object perception serves as a critical foundation for spatial reasoning. By first detecting objects in 3D space, models can reason more effectively over structured representations such as 3D bounding boxes, enhancing both the accuracy and interpretability of the reasoning process.

To this end, we propose \textbf{N3D-VLM}, a unified vision-language model that integrates 3D detection, grounding, and CoT reasoning. The model is equipped with inherent and generalizable 3D object perception, allowing it to accurately localize objects and capture depth cues in physical space. Building on these 3D perception results, N3D-VLM performs spatial reasoning tasks—such as computing inter-object distances based on 3D coordinates or inferring relative sizes from 3D bounding box dimensions, as shown in Fig.~\ref{fig:teaser}.
Recent works like~\cite{ma2025spatialreasoner, li2025spatialladder} utilized 3D coordinates or 2D grounding data to aid spatial understanding. However, our approach fundamentally differs: our model explicitly predicts comprehensive 3D bounding boxes, enabling more generalizable and interpretable spatial reasoning. 

A key challenge in developing general 3D object perception lies in the scarcity of large-scale 3D training data. Existing datasets~\cite{brazil2023omni3d,zhang2025detectany3d}, typically captured from indoor~\cite{baruch2021arkitscenes,roberts2021hypersim} or autonomous driving scenes~\cite{geiger2013kitti,caesar2020nuscenes}, suffer from limited diversity, small scale, and narrow category coverage. In contrast, large-scale 2D detection datasets~\cite{shao2019objects365,kuznetsova2020openimages,lin2014microsoftcoco} provide richer scene variety and class diversity.

To overcome 3D data scarcity, we leverage a depth and camera estimation model~\cite{wang2025moge} to lift 2D annotations into 3D space, generating abundant 3D detection and grounding data. We further construct spatial QA datasets to supervise CoT-based spatial reasoning. Another challenge is ensuring consistent real-world scale and camera geometry in 3D outputs; we address this by using the same depth model during data construction and introducing a depth-aware positional encoding to inject explicit depth cues. Jointly training on these heterogeneous sources enables our model to achieve robust 3D localization and effective CoT spatial reasoning grounded in explicit 3D perception.

In summary, we present a comprehensive solution for 3D grounded reasoning by introducing a model with native 3D obejct localization capabilities, marking a significant advancement in 3D visual understanding. Our main contributions are threefold: 

\begin{itemize} 
\item We propose a unified model that takes RGB-D input and features native 3D object detection/grounding abilities, enabling spatial reasoning based on detection outcomes. 
\item We design a data construction pipeline that leverages depth estimation to convert large-scale 2D annotations into 3D space, addressing the shortage of diverse and high-volume 3D object perception training data. 
\item We establish a spatial reasoning benchmark with explicit reasoning process, covering both single-object and multi-object scenarios across a wide variety of question types. 
\end{itemize}
Our model consistently surpasses existing methods in both object localization and spatial reasoning metrics, demonstrating state-of-the-art effectiveness and generalization. 
We will release code, checkpoints, and datasets upon acceptance.

%% file: 2_related.tex
\section{Related Work}
\label{sec:related_works}

\subsection{VLM for 3D Spatial Understanding}

Recently, an increasing number of approaches aim to extend general-purpose VLMs~\cite{hurst2024gpt,team2024gemini,liu2023llava,bai2025qwen2} with 3D spatial understanding, by leveraging 3D point clouds, video~\cite{chen2024ll3da,zheng2025video3dllm,qi2025gpt4scene,yang2025thinking,wang2025ross3d} or RGB/RGB-D inputs~\cite{chen2024spatialvlm, cheng2024spatialrgpt,he2025spatialormllm,zhu2024llava}. 
For example, GPT4Scene~\cite{qi2025gpt4scene} generates object-marked images from point clouds, and uses them along  BEV (bird's-eye view) images, to perform 3D captioning and question answering in reconstructed scenes. Think-in-Space~\cite{yang2025thinking} supports a broader range of spatial questions, including multiple-choice route planning and relative distance reasoning based on video inputs. However, these methods often rely on additional 3D information or object-level annotations, and are typically constrained to limited indoor environments. 
Other approaches focus on inferring spatial relations from 2D inputs.
SpatialVLM~\cite{chen2024spatialvlm} enables spatial question answering (e.g., left, right, front, behind) from a single RGB image. SpatialRGPT further extends this capability to region-level reasoning, allowing spatial understanding between specified image regions. Nevertheless, these methods generally lack explicit 3D spatial understanding, which either rely on black-box end-to-end reasoning to produce answers, or requires external modules for region localization. 
 
\subsection{VLM for 3D Object Localization}
Other 3D visual grounding approaches can localize object positions in 3D space based on point cloud or video input~\cite{li2025seeground,xu2024vlmgrounder,liu20233daxiesprompts,liu2025reasongrounder}. For example, VLM-Grounder~\cite{xu2024vlmgrounder} performs 2D segmentation on selected video frames, followed by multi-view matching and ensemble projection to localize objects in 3D. SeeGround~\cite{li2025seeground} supports 3D visual grounding but assumes known object positions after selecting a specific view. These methods typically rely on external segmentation tools or require additional object-level information. 
In contrast, our approach explicitly localizes 3D objects and outputs 3D bounding boxes. SpatialLM~\cite{mao2025spatiallm} also performs 3D grounding from point clouds and produces 3D bounding boxes. However, it is limited to indoor scenes with a small set of object categories and lacks spatial reasoning capabilities beyond grounding. 
SpatialReasoner~\cite{ma2025spatialreasoner} enhances spatial understanding by estimating object position and orientation. However, it operates in constrained scenarios, does not capture object size, and demonstrates limited generalization. By contrast, our method generalizes well across diverse scenes and outputs full 3D bounding boxes with complete object dimensions and positions, supporting both 3D detection, 3D grounding and downstream spatial reasoning.

%% file: 3_method.tex
\section{The Proposed Framework: N3D-VLM}
\label{sec:method}

\begin{figure*}[!tp]
  \centering   
  \includegraphics[width=1.0\linewidth]{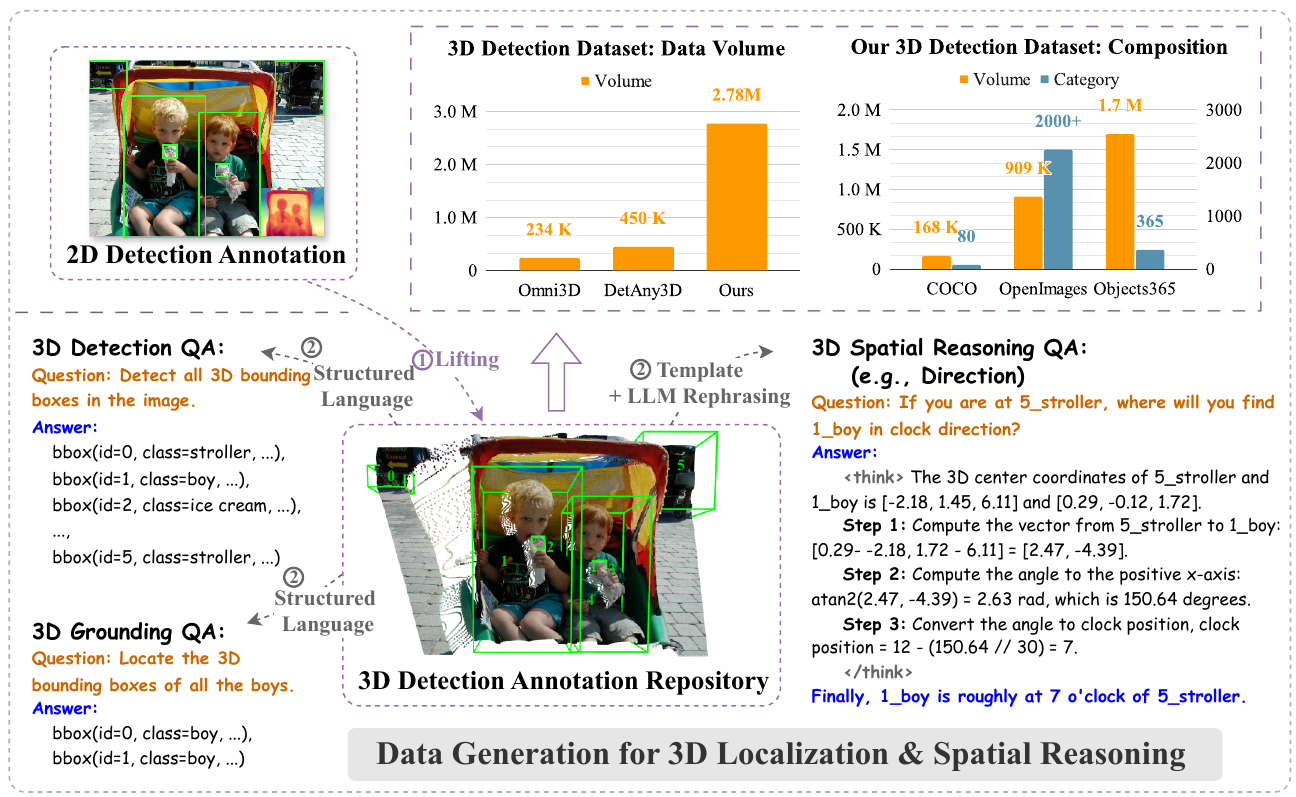}
   \vspace{-15pt}
   \caption{\textbf{Illustration of our data construction pipeline. }We first lift annotations from existing 2D detection datasets with diverse object categories into 3D space, resulting in a large-scale and category-rich 3D detection annotation repository. Based on this repository, we generate data for 3D detection, 3D grounding, and 3D spatial reasoning QA tasks. }
   \label{fig:data}
   \vspace{-10pt}
\end{figure*}

Given an RGB image and its corresponding monocular depth map, represented as $(I, D)$, we aim to train a vision-language model (VLM) that takes RGB-D input and outputs 3D object detection and grounding results in the form of explicit 3D bounding box in the camera coordinate system. Furthermore, the model can leverage these grounded 3D objects to perform spatial reasoning and answer spatial understanding questions. The depth map $D$ can be easily obtained using monocular depth estimation models~\cite{yang2024depthanythingv2,wang2025moge}. 

Our framework consists of two main components: 3D data construction and model design with training and evaluation, as illustrated in Fig.~\ref{fig:data} and Fig.~\ref{fig:pipeline}, respectively. 
As shown in Fig.~\ref{fig:data}, we begin by lifting 2D annotations from existing large-scale, category-rich 2D detection and grounding datasets into 3D domain, forming a diverse repository of 3D detection annotations. Based on this 3D repository, we further construct datasets for 3D detection, 3D grounding, and 3D spatial reasoning QA, which are used for both training and evaluation.
Fig.~\ref{fig:pipeline}a depicts the architecture of our model, which accepts RGB-D inputs. For grounding tasks, the model predicts structured language descriptions that correspond explicitly to the 3D bounding boxes of objects. For spatial reasoning tasks, the model performs spatial reasoning explicitly over grounded 3D objects to answer spatial understanding questions. 

\subsection{3D Data Generation}
\label{sec:3.1}
\noindent\textbf{3D Detection Annotation Repository. }
As shown in Fig.~\ref{fig:data}, we construct a large-scale 3D detection repository by lifting existing 2D detection datasets into 3D. 
Starting from images annotated with 2D bounding boxes, we first obtain object segmentation masks
~\cite{ravi2024sam} and estimate monocular depth for each image~\cite{wang2025moge}. The resulting depth maps are back-projected to generate point clouds in the camera space. By combining these point clouds with category and segmentation labels, we derive 3D bounding boxes for each object. During this lifting process, we apply rule-based filters to automatically remove outlier points from invalid depth values and discard implausible boxes that are excessively large or small. 
Leveraging the scale and category richness of 2D datasets, our constructed 3D corpus inherits these advantages, and provides a strong foundation for boosting VLMs' 3D localization capabilities. As illustrated in Fig.~\ref{fig:data}b, our dataset contains 2.78 million samples, which substantially more than existing single-image 3D detection datasets such as Omni3D ($\sim$234K)~\cite{brazil2023omni3d} and DetAny3D ($\sim$450K)~\cite{zhang2025detectany3d}.
Our dataset is constructed from three major sources: COCO~\cite{lin2014mscoco}, OpenImages~\cite{kuznetsova2020openimages}, and Objects365~\cite{shao2019objects365}. COCO and Objects365 offer high-quality 2D annotations across dozens to hundreds of object categories. For OpenImages, where the average number of boxes per image is low, we apply RAM~\cite{zhang2024recognize} to re-detect objects, producing open-vocabulary 2D detections that are subsequently lifted into 3D.

\begin{figure*}[!tp]
  \centering   \includegraphics[width=1.0\linewidth]{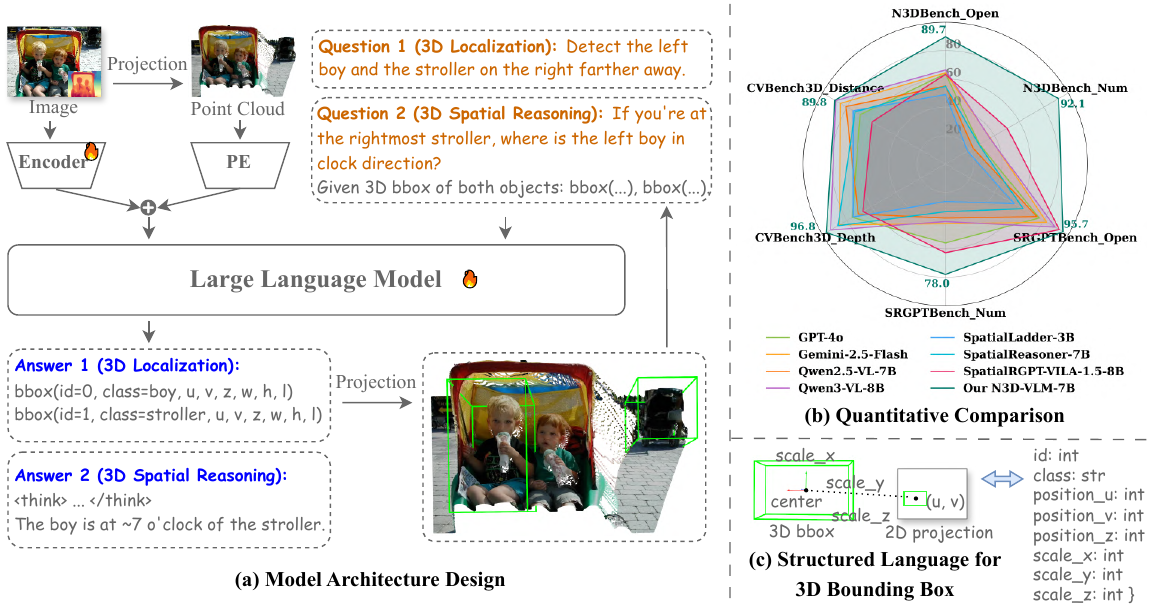}
   \vspace{-17pt}
   \caption{\textbf{Illustration of our model design and quantitative comparison.} (a) Overview of our model architecture and the cascaded spatial reasoning process. (b) Quantitative comparison showing that our model outperforms existing methods. (c) Definition of structured language representation for 3D bounding boxes. }
   \label{fig:pipeline}
   \vspace{-10pt}
\end{figure*}

\noindent\textbf{Generation of 3D Localization Data. }
After constructing the 3D detection annotation repository, we further generate corresponding 3D detection and grounding QA pairs to facilitate training VLMs for 3D object localization. 
Specifically, we represent each 3D bounding box using a structured language format, shown in Fig.~\ref{fig:pipeline}c. Similar to ~\cite{avetisyan2024scenescript,mao2025spatiallm}, 
each box is encoded as:
\vspace{-2mm}
\[
\mathrm{bbox}(id, class, u, v, z, s_x, s_y, s_z)
\]
where $id$ and $class$ denotes the object identifier and category. $(u,v)$ is the 2D projection of the 3D center on the image plane, and $z$ is its depth. 
$s\_x$, $s\_y$, and $s\_z$ represent the box dimensions along the three spatial axes. 
Note that, given known camera intrinsics, \((u, v, z)\) and \((x, y, z)\) are interconvertible via a deterministic projection.
For 3D detection QA, answers are directly derived from the structured annotations. 
For 3D grounding QA, we adopt two strategies. 
First, we select objects uniquely identifiable by category (i.e., categories that appear only once) and convert them into grounding questions. This extends to cases where multiple distinct categories are each uniquely identifiable. 
When multiple instances of the same category appear (see Fig.~\ref{fig:data}), we formulate questions like “Locate all the boys in the image.”
For objects not easily described by category, we either use referring expressions from existing datasets~\cite{kazemzadeh-etal-2014-referitgame}, or refer to them by rendering their 2D bounding boxes on the image. 
This process yields a diverse set of 3D detection and grounding QA pairs, supporting effective training of VLMs for 3D object localization. 

\noindent\textbf{Generation of 3D Spatial Reasoning Data. }
We further construct 3D spatial reasoning questions and explicit reasoning answers based on the 3D detection annotation repository. As shown in Fig.~\ref{fig:data}, we randomly sample objects from an image and apply predefined question templates, 
e.g., asking for the clock direction between two objects. For each question, we generate a deterministic reasoning process and answer based on the 3D bounding boxes, which is then rephrased using an LLM to improve naturalness. 
Following~\cite{cheng2024spatialrgpt}, we design both open-ended and numerically grounded questions. Specifically, we adopt question types similar to SpatialRGPT, including comparisons of relative scale (e.g., wider/narrower, taller/shorter), spatial relations (e.g., above/below, left/right, front/behind), absolute distances between objects, clock directions, and object dimensions (e.g., height, width).
In addition, we also extend to multi-object reasoning involving three or more objects. For example, we ask about relative distances among three objects or spatial configurations among a dozen objects, as illustrated in Fig.~\ref{fig:teaser}. 
All answers include deterministic numerical computations and interpretable reasoning steps, grounded in 3D object bounding boxes. 
In the example from Fig.~\ref{fig:data}, the question asks for the clock direction of the boy relative to the stroller. The reasoning process is first generated based on a predefined template: given the 3D bounding boxes of both objects, we compute the vector from the stroller to the boy on the $xz$-plane, calculate its angle with respect to the positive $x$-axis, and convert the angle into a clock position. 
This explicit geometric reasoning is embedded in the answer chain, enabling more interpretable and intuitive explanations.
More question templates and task definitions are provided in the Appendix.

\subsection{Model Architecture}
\label{sec:3.2}
\noindent\textbf{3D-aware Visual Encoding. }
To ensure that the 3D bounding boxes predicted by our VLM are in real-world scale and aligned with an existing coordinate system, we adopt the depth estimation model~\cite{wang2025moge}, which predicts both depth maps and camera intrinsics. 
We then use the predicted depth as an additional input to our model. 
This guarantees that all predicted 3D bounding boxes are expressed in metric scale and aligned with the coordinate system defined by~\cite{wang2025moge}. 

As illustrated in Fig.~\ref{fig:pipeline}, we design a 3D-aware visual encoding pipeline to incorporate geometric information into the vision-language model. Given an RGB image \( I \in \mathbb{R}^{H \times W \times 3} \), its corresponding depth map \( D \in \mathbb{R}^{H \times W} \), and camera intrinsics \( \mathrm{intr} \in \mathbb{R}^{3 \times 3} \). We first back-project each pixel to a 3D point in the camera coordinate system: \( P \in \mathbb{R}^{H \times W \times 3} \):
\begin{equation}
P_{ij} = D_{ij} \cdot \mathrm{intr}^{-1} \cdot 
\begin{bmatrix}
u_j \\
v_i \\
1
\end{bmatrix},
\label{eq:backprojection}
\end{equation}
where \( (u_j, v_i) \) are the pixel coordinates. This yields a dense point cloud \( P \in \mathbb{R}^{H \times W \times 3} \), which is then downsampled to \( \hat{P} \in \mathbb{R}^{h \times w \times 3} \) to match the spatial resolution of the image features \( F_{\text{img}} \in \mathbb{R}^{h \times w \times c} \) extracted by the vision encoder.
To inject spatial information, each 3D coordinate \( (x, y, z) \in \hat{P} \) is encoded using sinusoidal positional encoding. For each axis \( \alpha \in \{x, y, z\} \), we compute:
\begin{align}
\mathrm{PE}(\alpha, 2i) &= \sin\left(\frac{\alpha}{10000^{2i / c}}\right), \notag \\
\mathrm{PE}(\alpha, 2i+1) &= \cos\left(\frac{\alpha}{10000^{2i / c}}\right),
\label{eq:pe}
\end{align}
for \( i = 0, 1, \dots, \frac{c}{2} - 1 \). The final coordinate embedding is obtained by summing the encodings across all three axes:
\begin{equation}
\mathbf{e}^{\text{coord}} = \sum_{k \in \{x, y, z\}} \mathrm{PE}(k).
\label{eq:coord_sum}
\end{equation}
We then add the coordinate embedding to the image features to obtain the fused representation as follows:
\begin{equation}
\tilde{F}_{\text{img}} = F_{\text{img}} + \mathbf{e}^{\text{coord}}.
\label{eq:fusion}
\end{equation}
The fused feature map \( \tilde{F}_{\text{img}} \), which encodes both visual and spatial cues, is then passed to the language model along with the prompt tokens for autoregressive prediction.

\noindent\textbf{Training Strategy and Inference Pipeline. }
Our model is based on Qwen2.5-VL~\cite{bai2025qwen2} and trained in two stages. In the first stage, we train the model for 3D object localization using the dataset described in Sec.~\ref{sec:3.1}. In the second stage, we train the model for grounding-based 3D spatial reasoning using a mixture of 3D spatial reasoning data and a subset of the localization data. All parameters of the encoder and the language model are learnable throughout both stages. 

At inference time, our model supports two usage modes. The first, illustrated in Fig.~\ref{fig:teaser}, allows users to ask spatial-related questions directly. The model automatically decomposes the query into two steps: 3D object grounding, followed by spatial reasoning based on the grounding results. 
In the second mode, shown in Fig.~\ref{fig:pipeline}, users can explicitly request 3D grounding first, then issue follow-up questions according to the grounding output. In both cases, reasoning is performed conditionally on the grounding results. 

\subsection{N3D-Bench}
Given the narrow range of scenes and object categories, current benchmarks fall short in representing the complexity of real-world 3D spatial understanding. 
Based on our 3D spatial data generation pipeline, we manually curated 1,200 open-ended and 800 numerically questions with CoT reasoning to construct N3D-Bench. 
As shown in Tab.~\ref{tab:bench_comparison}, our benchmark significantly extends SpatialRGPT-Bench~\ref{tab:bench_comparison} in both object category coverage and question complexity. N3D-Bench includes references to 264 object categories, which is three times more than SpatialRGPT-Bench. It also introduces questions involving spatial relations among three or more objects, as well as viewpoint-shifted reasoning (e.g., “from the opposite view”), which are not considered in SpatialRGPT-Bench. Additionally, we introduce explicit CoT reasoning grounded in 3D object bounding box, offering interpretable intermediate steps beyond direct answers. 
These enhancements make N3D-Bench a more comprehensive and challenging benchmark for evaluating 3D spatial reasoning over both single and multiple objects. 

\begin{table}[!t]
\centering
\caption{Comparison between our proposed N3D-Bench and SpatialRGPT-Bench~\cite{cheng2024spatialrgpt}. }
\vspace{-6pt}
\renewcommand{\arraystretch}{1.1}
\newcommand{\cmark}{\textcolor{green!90!black}{\ding{51}}} 
\newcommand{\xmark}{\textcolor{red!90!black}{\ding{55}}}   
\resizebox{\linewidth}{!}{
\begin{tabular}{l|cc}
\toprule
\textbf{Comparision} & \textbf{SpatialRGPT-Bench~\cite{cheng2024spatialrgpt}} & \textbf{N3D-Bench} \\
\midrule
\# Questions & 1406 & 2000 \\
\# Object Categories & 88 & 264 \\
Objects / Question & \{1,2\} & \{1,2,3,$>$3\} \\
View Change & \xmark & \cmark \\
CoT Reasoning & \xmark & \cmark \\
\bottomrule
\end{tabular}
}
\vspace{-10pt}
\label{tab:bench_comparison}
\end{table}

%% file: 4_exp.tex
\section{Experiments}
\label{sec:exp}

\begin{table*}[htbp]
  \centering
  \caption{\textbf{Quantitative comparison on spatial reasoning benchmarks.} Our N3D-VLM consistently outperforms baseline methods across all three spatial reasoning benchmarks, achieving state-of-the-art performance on open-ended, numerical, and multiple-choice questions. }
  \vspace{-6pt}
  \resizebox{\linewidth}{!}{
    \begin{tabular}{llcc|cc|c}
      \toprule
      \multirow{2}{*}{Category} & \multirow{2}{*}{Method} & \multicolumn{2}{c|}{N3D-Bench} & \multicolumn{2}{c|}{SpatialRGPT-Bench~\cite{cheng2024spatialrgpt}} & \multicolumn{1}{c}{CV-Bench-3D~\cite{tong2024cambrian1}} \\
      \cmidrule(lr){3-4} \cmidrule(lr){5-6} \cmidrule(lr){7-7}
        &  & Open-ended & Numerical & Open-ended & Numerical & Multi-choice \\
      \midrule
      \multirow{2}{*}{Closed-source} 
        & GPT-4o~\cite{hurst2024gpt} & 63.5 & 27.8 & 76.3 & 55.8 & 72.4 \\
        & Gemini-2.5-Flash~\cite{team2024gemini} & 64.2 & 36.7 & 82.4 & 42.2 & 86.0 \\
      \midrule
      \multirow{5}{*}{Open-source} 
        & Qwen2.5-VL-7B~\cite{bai2025qwen2} & 55.0 & 22.5 & 74.4 & 38.2 & 75.8 \\
        & Qwen3-VL-8B~\cite{Qwen3-VL} & 66.3 & 36.3 & 89.2 & 40.7 & 91.3 \\
        & SpatialLadder-3B~\cite{li2025spatialladder} & 48.9 & 18.1 & 55.9 & 26.5 & 74.9 \\
        & SpatialReasoner-7B~\cite{ma2025spatialreasoner} & 54.8 & 27.4 & 63.2 & 33.7 & 80.3 \\
        & SpatialRGPT-VILA-1.5-8B~\cite{cheng2024spatialrgpt} & 63.1 & 50.4 & 92.7 & 62.7 & 63.3 \\
      \midrule
      \multirow{2}{*}{Ours} 
        & N3D-VLM-3B & 77.0 & 90.1 & 80.5 & 73.3 & \textbf{96.3} \\
        & N3D-VLM-7B & \textbf{89.7} & \textbf{92.1} & \textbf{95.7} & \textbf{78.0} & 93.3 \\
      \bottomrule
    \end{tabular}
  }
  \vspace{-4pt}
  \label{tab:qa}
\end{table*}

\begin{table*}[htbp]
  \centering
  \caption{\textbf{Quantitative comparison of 3D grounding using IoU and center offset of projected 3D bounding boxes.} Our N3D-VLM achieves the best performance on both projected IoU and offset, demonstrating our superior accuracy in localizing objects in 3D space. }
  \vspace{-6pt}
  \resizebox{\linewidth}{!}{
    \begin{tabular}{lcc|cc|cc|cc}
      \toprule
      \multirow{2}{*}{Method} & \multicolumn{2}{c|}{Refcoco~\cite{kazemzadeh-etal-2014-referitgame}} & \multicolumn{2}{c|}{Refcoco+~\cite{kazemzadeh-etal-2014-referitgame}} & \multicolumn{2}{c|}{Refcocog~\cite{kazemzadeh-etal-2014-referitgame}} & \multicolumn{2}{c}{Objects365~\cite{shao2019objects365}} \\
      \cmidrule(lr){2-3} \cmidrule(lr){4-5} \cmidrule(lr){6-7} \cmidrule(lr){8-9}
       & Proj. IoU $\uparrow$ & Proj. Offset $\downarrow$ & Proj. IoU $\uparrow$ & Proj. Offset $\downarrow$ & Proj. IoU $\uparrow$ & Proj. Offset $\downarrow$ & Proj. IoU $\uparrow$ & Proj. Offset $\downarrow$ \\
      \midrule
        Qwen3-VL-8B~\cite{Qwen3-VL} & 0.37 & 0.16 & 0.34 & 0.26 & 0.36 & 0.14 & 0.28 & 0.12 \\
        Qwen3-VL-30B-A3B~\cite{Qwen3-VL} & 0.38 & 0.14 & 0.36 & 0.16 & 0.38 & 0.13 & 0.28 & 0.13\\
      \midrule
        Our N3D-VLM-7B & \textbf{0.59} & \textbf{0.06} & \textbf{0.53} & \textbf{0.10} & \textbf{0.54} & \textbf{0.08} & \textbf{0.61} & \textbf{0.05} \\
      \bottomrule
    \end{tabular}
  }
  \vspace{-10pt}
  \label{tab:grounding}
\end{table*}

\begin{table}[htbp]
  \centering
  \caption{\textbf{Quantitative comparison of 3D bounding boxes.} Our N3D-VLM achieves the best results across 3D IoU and 3D offset, demonstrating superior performance in 3D object localization. }
  \vspace{-6pt}
  \resizebox{0.75\linewidth}{!}{
    \begin{tabular}{lcc}
      \toprule
      \multirow{2}{*}{Method} & \multicolumn{2}{c}{Refcoco/+/g~\cite{kazemzadeh-etal-2014-referitgame}} \\
      \cmidrule(lr){2-3}
       & 3D IoU $\uparrow$ & 3D Offset $\downarrow$ \\
      \midrule
        Qwen3-VL-8B~\cite{Qwen3-VL} & 0.20 & 1.88 \\
        Qwen3-VL-30B-A3B~\cite{Qwen3-VL} & 0.27 & 1.86 \\
      \midrule
        Our N3D-VLM-7B & \textbf{0.48} & \textbf{0.36} \\
      \bottomrule
    \end{tabular}
  }
  \vspace{-8pt}
  \label{tab:grounding2}
\end{table}

\subsection{Experimental Setup}
\par\noindent\textbf{Dataset.}
As described in Sec.~\ref{sec:3.2}, we train our model on 3D object localization and spatial reasoning data derived from OpenImages, Objects365, and COCO. 
For 3D spatial reasoning, we evaluate on three benchmarks: our proposed N3D-Bench, SpatialRGPT-Bench~\cite{cheng2024spatialrgpt} (1,404 open-ended and numerical questions), and CV-Bench-3D~\cite{tong2024cambrian1} (1,200 multiple-choice questions). 
For 3D grounding, we evaluate on the RefCOCO series~\cite{kazemzadeh-etal-2014-referitgame} , along with an additional test set constructed from Objects365. 
\par\noindent\textbf{Metrics.}
For spatial reasoning, we report the accuracy. For open-ended questions, we use GPT-4o~\cite{hurst2024gpt} in an LLM-as-a-judge setup to access correctness. For numerical questions, we extract predicted values via string matching and apply a ±25\% tolerance, following~\cite{cheng2024spatialrgpt}. 
For all methods evaluated on SpatialRGPT-Bench, except SpatialRGPT itself, object references are provided via bounding boxes drawn on the image. 
For 3D grounding, to ensure fair comparison across varying depth scales and box types, we compute projected IoU and projected center offset by projecting predicted 3D bounding box onto the image plane and comparing it with ground-truth 2D boxes. We also try to align the predicted boxes to the depth of the ground-truth boxes and compute 3D IoU and 3D center offset. To mitigate the alignment noise, 3D metrics are reported on a sampled subset.

\subsection{Main Results}
\par\noindent\textbf{3D Spatial QA. }
Tab.~\ref{tab:qa} compares the accuracy of our method with baseline approaches across three benchmarks. Our model achieves the highest accuracy on all benchmarks, demonstrating that our native 3D grounding significantly improves performance on 3D spatial reasoning tasks. 
Compared to the base model Qwen2.5-VL-7B, our 7B model shows substantial gains, especially on numerical questions, which indicates that grounding-based reasoning enhances the model's quantitative understanding beyond standard QA capabilities. 
While Qwen3-VL improves over Qwen2.5-VL in spatial reasoning, its numerical reasoning remains limited, with only 36.3\% and 40.7\% accuracy. In contrast, our model achieves 92.1\% and 78.0\% on the same tasks, demonstrating significantly stronger numerical reasoning. 
Although SpatialRGPT performs well on numerical questions in N3D-Bench, achieving 50.4\% which outperforms Qwen3-VL, but still far below our model’s 92.1\%.

\par
\begin{figure*}[!tp]
    \centering
    \includegraphics[width=0.98\linewidth]{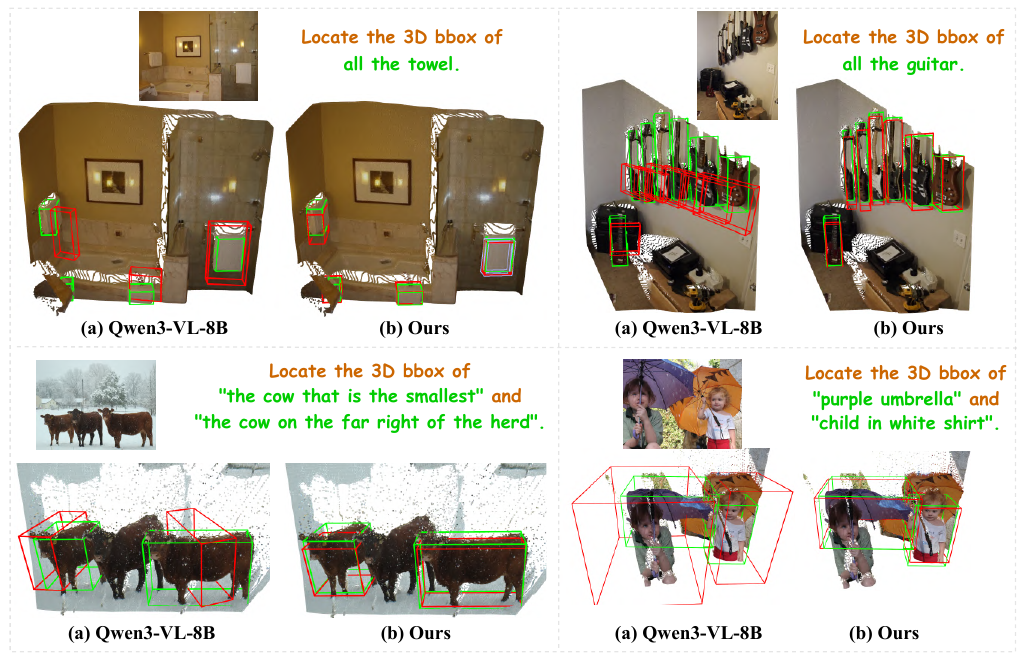}
    \vspace{-10pt}
    \caption{\textbf{Qualitative comparison of 3D grounding capability with Qwen3-VL-8B~\cite{Qwen3-VL}.} Compared to Qwen3-VL-8B, our N3D-VLM generates 3D bounding boxes that more accurately close to the ground truth, reflecting stronger 3D understanding and localization precision. In the visualization, \textcolor{green!90!black}{green boxes} represent ground truth 3D bounding boxes, and \textcolor{red!90!black}{red boxes} indicate model’s predictions. }
    \label{fig:grounding}
    \vspace{-10pt}
\end{figure*}

\noindent\textbf{3D Object Grounding.}~Tab.~\ref{tab:grounding}, Tab.~\ref{tab:grounding2} and Fig.~\ref{fig:grounding} present the quantitative and qualitative comparisons of 3D grounding performance between our model and Qwen3-VL~\cite{Qwen3-VL}. 
As shown in Tab.~\ref{tab:grounding} and ~\ref{tab:grounding2}, our method consistently outperforms Qwen3-VL in both projected 2D metrics and aligned 3D bounding box evaluation, demonstrating stronger 3D grounding capabilities. 
Fig.~\ref{fig:grounding} further shows that our model accurately localize objects in 3D space, producing precise 3D bounding boxes across diverse scenarios, including indoor and outdoor environments. 
We provide more grounding results comparison and analysis in the Appendix.

\begin{table}[!tp]
  \centering
  \caption{\textbf{Ablation studies on model design.} Variants (3) and (4), which adopt our model design shown in Fig.~\ref{fig:pipeline}(a), achieve the best 3D detection performance, when compared with other variants. }
  \vspace{-6pt}
  \resizebox{0.95\linewidth}{!}{
    \begin{tabular}{lccc}
      \toprule
      Method & F1@0.25 $\uparrow$ & P@0.25 $\uparrow$ & R@0.25 $\uparrow$ \\
      \midrule
        (0) SpatialLM~\cite{mao2025spatiallm}-340K & 2.2 & 2.3 & 2.4 \\
      \midrule
        (1) 3B-340K-nodepth & 9.4 & 10.9 & 9.4 \\
        (2) 3B-340K-cameraxy & 10.8 & 11.6 & 10.7 \\
      \midrule
        (3) 3B-340K-imageuv & 12.8 & 13.6 & 12.9 \\
        (4) 3B-1.7M-imageuv & \textbf{22.9} & \textbf{24.3} & \textbf{22.9} \\
      \bottomrule
    \end{tabular}
  }
  \vspace{-4pt}
  \label{tab:ablation_detection}
\end{table}

\subsection{Ablation Study on Model Design}
We conduct ablation studies of the 3D detection task on Objects365 dataset to evaluate the effectiveness of our model design. Tab.~\ref{tab:ablation_detection} reports the 3D IoU results on a validation set of 5,565 images covering 341 object classes. 
Row (0) shows that the SpatialLM~\cite{mao2025spatiallm} architecture, which combines a point cloud encoder~\cite{wu2025sonata} and an LLM~\cite{bai2025qwen2}, performs poorly on our more diverse point cloud data, indicating limited generalization. 
Variants (1) and (2) are based on our model architecture, with (1) removing depth input and (2) using depth but directly predicting 3D coordinates in the camera coordinate system. 
Variants (3) and (4) follow our full design (Fig.~\ref{fig:pipeline}a), which takes depth input and predicts 2D pixel-space coordinates shown in Fig.~\ref{fig:pipeline}c.
Comparing (1) and (3), we can observe that incorporating depth input improves 3D detection accuracy, increasing the F1 score from 9.4 to 12.8. 
Comparing (2) and (3), we can find that predicting the center coordinates in pixel space outperforms camera-space prediction, likely because the base model is pretrained with 2D perception data, which is naturally aligned with the image-space representations. 
These results suggest that using depth input and predicting pixel-space coordinates, is more effective than excluding depth or directly predicting in the camera coordinate system. 
Finally, scaling the training set from 340K to 1.7M samples in (4) leads to substantial improvements, demonstrating the effectiveness of our large-scale 3D data generation pipeline. 

\begin{table}[!tp]
  \centering
  \caption{\textbf{Ablation studies on the effectiveness of 3D grounding.} Our intermediate 3D grounding results can improve Qwen3-VL’s performance, while training our model directly on QA data leads to degraded results. These findings demonstrate that our unified N3D-VLM, which first performing grounding, then spatial reasoning, effectively enhances the spatial understanding. }
  \vspace{-6pt}
  \resizebox{0.9\linewidth}{!}{
    \begin{tabular}{lcc}
      \toprule
      Method & N3D-open $\uparrow$ & N3D-num $\uparrow$ \\
     \midrule
        Qwen3-VL-8B~\cite{Qwen3-VL} (direct answer) & 66.3 & 36.3 \\
        Qwen3-VL-8B~\cite{Qwen3-VL} (ground. given) & 71.3 & 54.6 \\
        \textit{Improvement (\%)} & \textit{+7.5\%} & \textit{+50.4\%} \\
      \midrule
        Our QAonly-7B  & 80.6 & 62.4 \\
        Our N3D-VLM-7B & \textbf{89.7} & \textbf{92.1} \\
      \bottomrule
    \end{tabular}
  }
  \vspace{-10pt}
  \label{tab:ablation_qa}
\end{table}

\subsection{3D Grounding Helps Spatial Reasoning}
We design two experiments to demonstrate that native 3D grounding explicitly improves spatial reasoning. 
First, we feed the grounding results from our model into Qwen3-VL, prompting it to reason based on our 3D grounding output. As shown in Tab.~\ref{tab:ablation_qa}, Qwen3-VL achieves higher accuracy with the grounding results compared to answering directly, indicating that 3D grounding enhances spatial reasoning. 
However, its performance still lags behind our 7B model, suggesting that our model performs strong spatial reasoning based on 3D grounding. 
In the second experiment, we use the same architecture but train it end-to-end for question answering without separating grounding and reasoning. This setup underperforms our full model, confirming that explicitly decomposing the task into grounding and reasoning leads to better performance under same architecture. 

%% file: 5_conclusion.tex
\section{Conclusion}
\label{sec:conclusion}
We present a unified framework N3D-VLM that bridges 3D object perception and spatial reasoning within a single model. By enabling native 3D grounding and explicit 3D-aware reasoning, our approach offers both accurate localization and interpretable spatial understanding. 
Supported by a scalable data construction pipeline that projects 2D annotations into 3D space and facilitates the creation of explicit reasoning datasets, our model demonstrates reasonable generalization in 3D grounding and provides a structured foundation for spatial reasoning. 
Extensive experiments validate the effectiveness of our framework, which achieves strong performance in both 3D grounding and 3D spatial reasoning among existing vision-language models. 

%% file: X_suppl.tex
\setcounter{page}{1}
\maketitlesupplementary

\appendix
In this supplementary material, we provide more additional experiments in Sec.~\ref{sec:supple_exp}. 
We also present a video demo for more qualitative results in Sec.~\ref{sec:supple_video}. 
We illustrate the data distribution of our proposed N3D-Bench in Sec.~\ref{sec:supple_n3d}.

\section{Additional Experiments}
\label{sec:supple_exp}

\subsection{3D Grounding Comparison}
\begin{figure*}[!tp]
    \centering
    \includegraphics[width=1.0\linewidth]{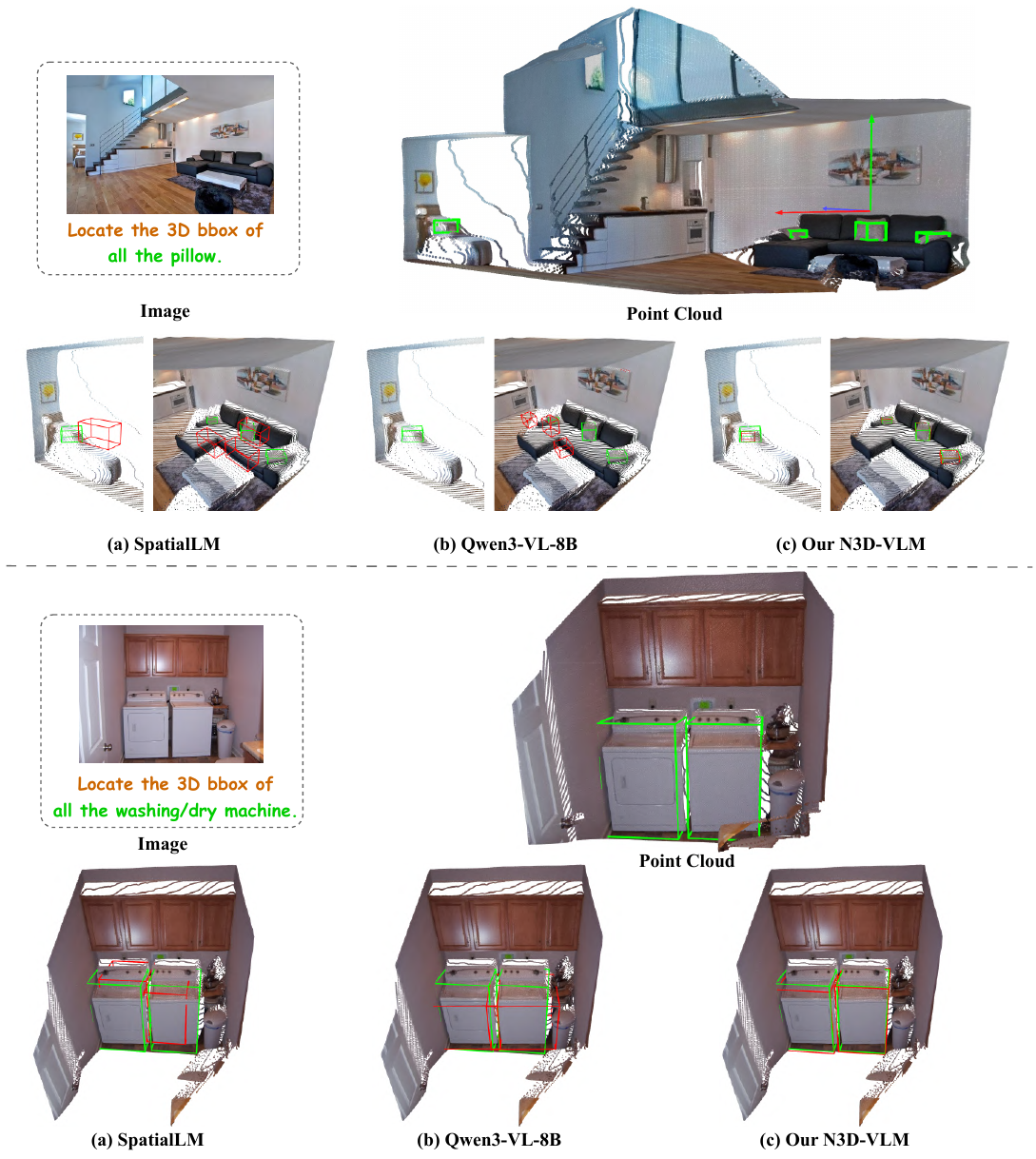}
    \vspace{-10pt}
    \caption{\textbf{Qualitative comparison of 3D grounding capability with SpatialLM~\cite{mao2025spatiallm} and Qwen3-VL-8B~\cite{Qwen3-VL} in indoor scenes.}  our N3D-VLM accurately localizes objects such as pillows and washing machines, while baselines either miss objects or exhibit inaccurate prediction. In the visualization, \textcolor{green!90!black}{green boxes} represent ground truth 3D bounding boxes, and \textcolor{red!90!black}{red boxes} indicate model’s predictions. }
    \label{fig:supple_grounding_a}
    \vspace{-10pt}
    % \vspace{-18pt}
\end{figure*}

\begin{figure*}[!tp]
    \centering
    \includegraphics[width=1.0\linewidth]{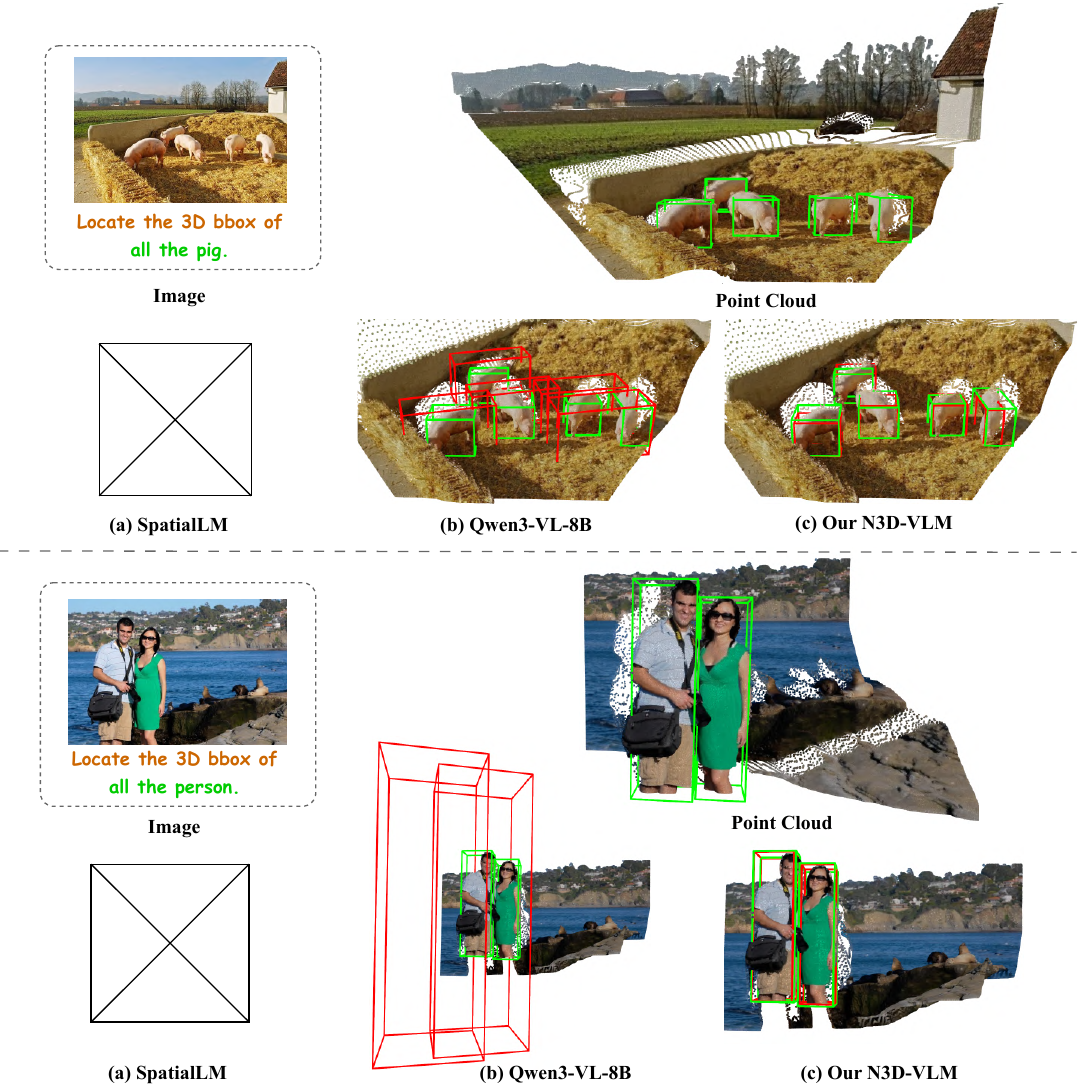}
    \vspace{-10pt}
    \caption{\textbf{Qualitative comparison of 3D grounding capability with SpatialLM~\cite{mao2025spatiallm} and Qwen3-VL-8B~\cite{Qwen3-VL} in outdoor scenes.} Our N3D-VLM outperforms Qwen3-VL-8B and SpatialLM, highlighting its superior native 3D grounding capability for reliable spatial reasoning. In the visualization, \textcolor{green!90!black}{green boxes} represent ground truth 3D bounding boxes, and \textcolor{red!90!black}{red boxes} indicate model’s predictions. }
    \label{fig:supple_grounding_b}
    \vspace{-10pt}
    % \vspace{-18pt}
\end{figure*}

We present a qualitative comparison of the native 3D grounding capability of our N3D-VLM with two baseline methods, SpatialLM~\cite{mao2025spatiallm} and Qwen3-VL-8B~\cite{Qwen3-VL}, as shown in Fig.~\ref{fig:supple_grounding_a} and Fig.~\ref{fig:supple_grounding_b}.
% how
SpatialLM is a vision-language model designed for 3D grounding in the indoor scenes using point cloud input. We generate point clouds for SpatialLM by combining the input image with depth maps obtained via ~\cite{wang2025moge}. For Qwen3-VL-8B~\cite{Qwen3-VL}, we apply the depth alignment strategy described in the main paper to align its predictions with our ground truth. 
% analysis
Fig.~\ref{fig:supple_grounding_a} shows two indoor scene examples. In the first, our method accurately localizes the pillows, while both baselines either detect only a subset of objects or exhibit inaccurate spatial prediction. In the second scene, our method also provides a precise localization of the washing machines.
% analysis
Fig.~\ref{fig:supple_grounding_b} presents two diverse outdoor scenes. SpatialLM fails in these cases, as it is limited to a predefined set of indoor categories. In both examples, our N3D-VLM outperforms Qwen3VL-8B for detection, demonstrating superior native 3D grounding capabilities, which are essential for reliable spatial reasoning.

% spatialreasoner
We also consider SpatialReasoner~\cite{ma2025spatialreasoner}, a related method capable of predicting 3D object centers. However, it does not support explicit 3D bounding box output, limiting its capacity for comprehensive spatial perception. In our evaluation, we observed that even when prompted to output object centers, the results were inconsistent and often failed to follow a coherent 3D coordinate format.

\subsection{3D Spatial Reasoning Comparison}
\begin{figure*}[!tp]
    \centering
    \includegraphics[width=0.9\linewidth]{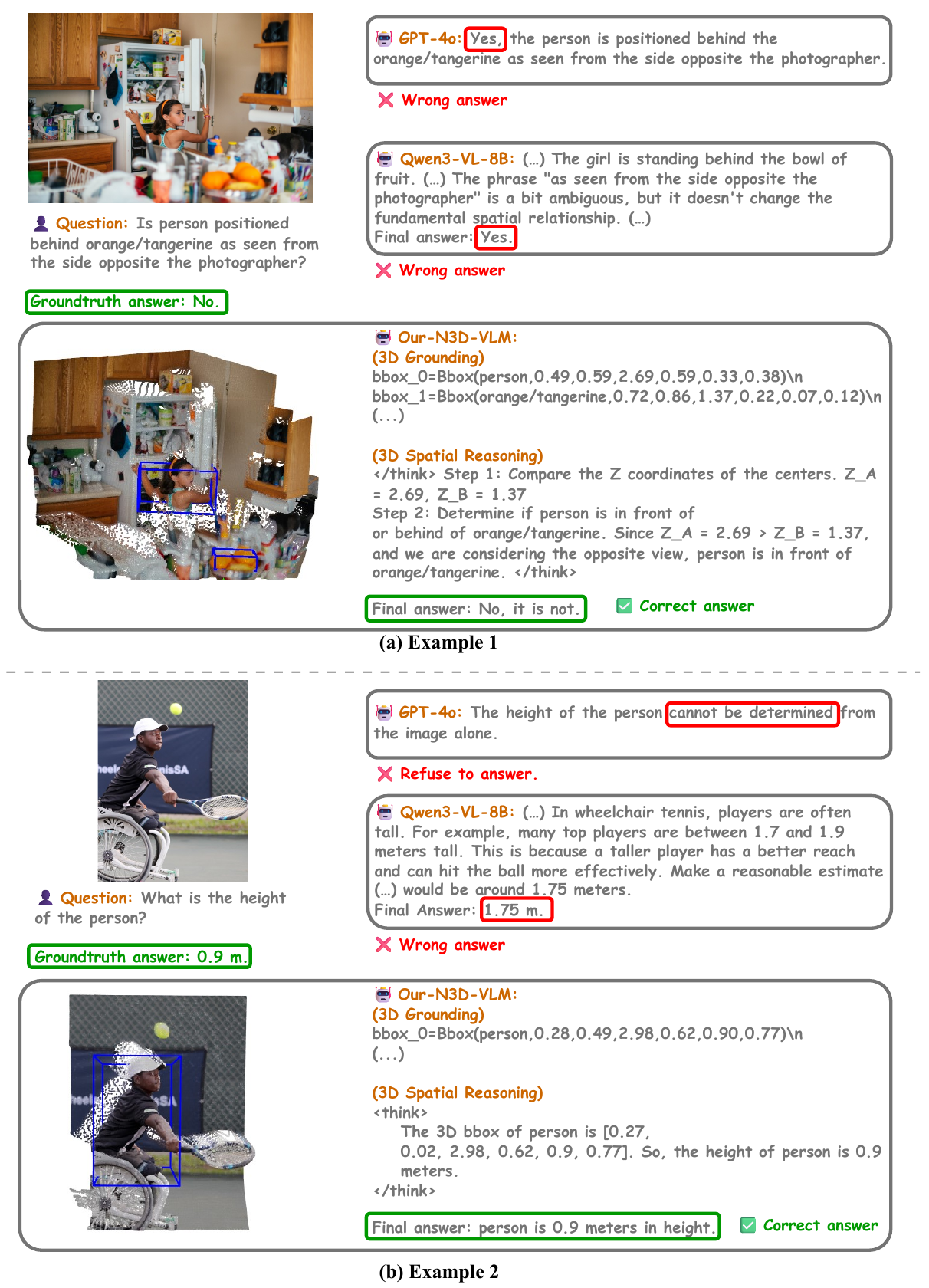}
    \vspace{-10pt}
    \caption{\textbf{Qualitative comparison of 3D spatial reasoning capability with GPT-4o~\cite{hurst2024gpt} and Qwen3-VL-8B~\cite{Qwen3-VL}.} Our N3D-VLM outperforms GPT-4o and Qwen3-VL-8B by leveraging accurate 3D grounding and 3D spatial reasoning.}
    \label{fig:supple_reasoning_a}
    \vspace{-10pt}
    % \vspace{-18pt}
\end{figure*}

\begin{figure*}[!tp]
    \centering
    \includegraphics[width=0.9\linewidth]{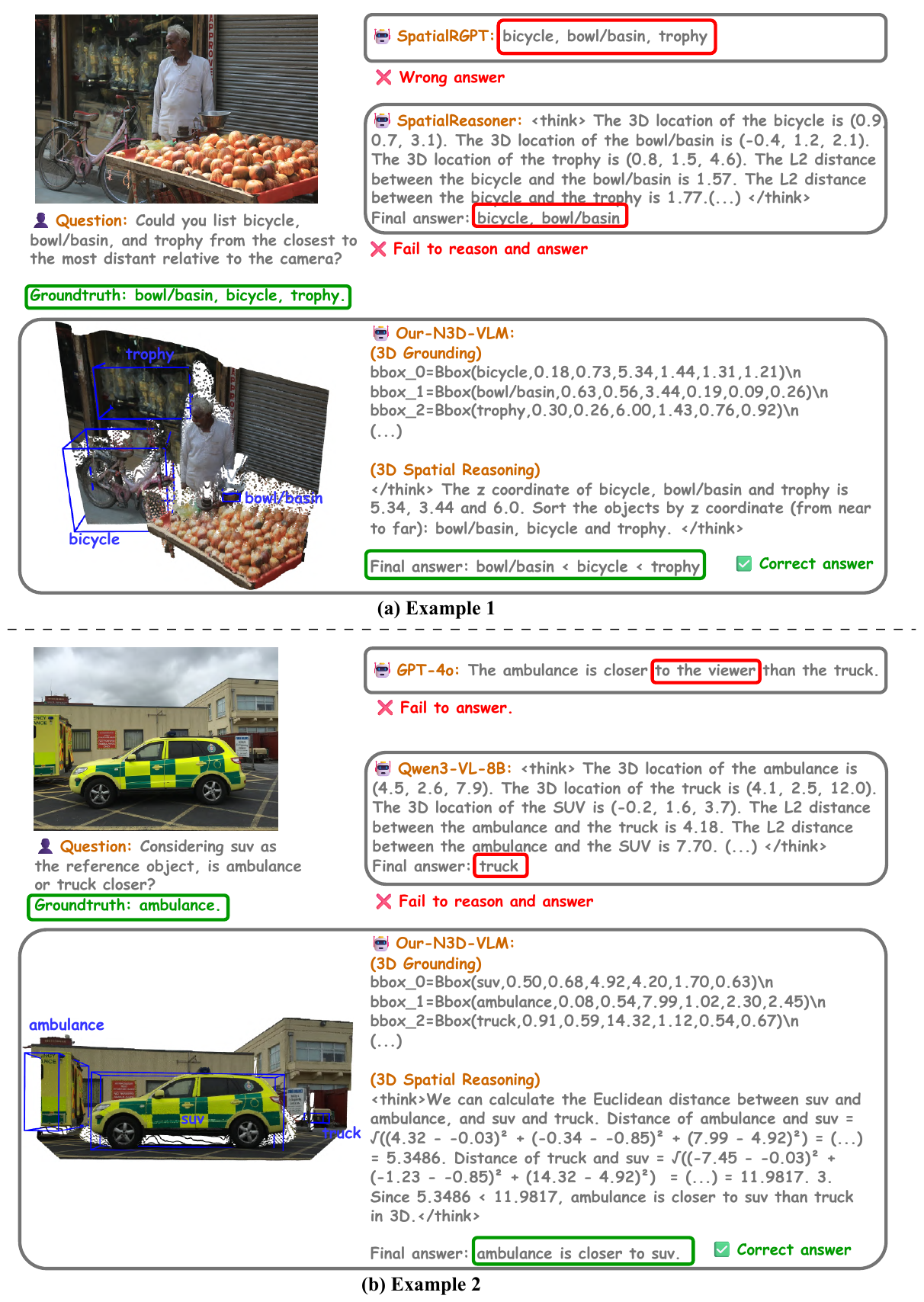}
    \vspace{-10pt}
    \caption{\textbf{Qualitative comparison of 3D spatial reasoning capability with SpatialRGPT~\cite{cheng2024spatialrgpt} and SpatialReasoner~\cite{ma2025spatialreasoner}.} N3D-VLM outperforms SpatialRGPT and SpatialReasoner by accurately interpreting the question and reasoning over explicit 3D bounding boxes. }
    \label{fig:supple_reasoning_b}
    \vspace{-10pt}
    % \vspace{-18pt}
\end{figure*}

% Done: tab.a/b, sub-score of n3d/spatialrgpt bench
% P0.5: figure, 3+ examples in video demo. 
We present a qualitative comparison of the 3D spatial reasoning capabilities of our N3D-VLM with GPT-4o~\cite{hurst2024gpt}, Qwen3-VL-8B~\cite{Qwen3-VL}, SpatialRGPT~\cite{cheng2024spatialrgpt}, and SpatialReasoner~\cite{ma2025spatialreasoner}, as shown in Fig.~\ref{fig:supple_reasoning_a} and Fig.~\ref{fig:supple_reasoning_b}.
% analysis
Fig.~\ref{fig:supple_reasoning_a} compares our N3D-VLM with GPT-4o and Qwen3-VL-8B. In Example 1, both GPT-4o and Qwen3-VL fail to answer the question correctly due to incorrect reasoning about the change in viewpoint. In contrast, our N3D-VLM provides the correct answer by leveraging accurate 3D grounding and reasoning over the grounded results. In Example 2, GPT-4o and Qwen3-VL-8B overly rely on commonsense priors while neglecting visual cues. Our method, empowered by native 3D grounding, directly localizes the bounding box to answer the question accurately.
% analysis
Fig.~\ref{fig:supple_reasoning_b} compares our method with SpatialRGPT and SpatialReasoner on questions involving relative distance comparison and depth comparison among three objects. SpatialRGPT either provides incorrect answers or misunderstands the question entirely. Although SpatialReasoner has the potential to reason using 3D coordinates, it fails to complete the reasoning process. For instance, in Example 1, it incorrectly attempts to calculate absolute distances for a depth-ordering question, resulting in rigid and ultimately incorrect reasoning. 

In contrast, our N3D-VLM not only accurately localizes 3D bounding boxes of objects but also reliably performs reasoning over grounded results, demonstrating its effectiveness in 3D spatial understanding tasks.

% sub-score
As a supplement to Table 2 in the main paper, Tab.~\ref{tab:supple_n3d} and Tab.~\ref{tab:supple_spatialrgptbench} report accuracy scores across different question types. Our N3D-VLM consistently outperforms our base model Qwen2.5VL~\cite{bai2025qwen2}, confirming that native 3D grounding significantly enhances the model’s ability to understand various forms of spatial questions. 

\begin{table*}[!tp]
  \centering
  \caption{\textbf{Detailed quantitative comparison on spatial reasoning benchmark N3D-Bench.} \cellcolor{best}{Dark green} indicates the highest accuracy, while \cellcolor{second}{light green} denotes the second-highest accuracy.}
  \vspace{-6pt}
  \resizebox{\linewidth}{!}{
    \begin{tabular}{llccccccccccc}
      \toprule
      \multirow{3}{*}{Category} & \multirow{3}{*}{Method} & \multicolumn{11}{c}{N3D-Bench} \\
      \cmidrule(lr){3-13}
      % Category & Method & Open-ended & Numerical & Open-ended & Numerical & Multi-choice \\
        &  & \multicolumn{7}{c}{Open-ended} & \multicolumn{4}{c}{Numerical} \\
        \cmidrule(lr){3-9} \cmidrule(lr){10-13}
        &  & left/right & front/behind & wide/thin & tall/short & big/small & relative dis. & depth comp. & width/height & distance & ver./hor. dis. & direction \\
      \midrule
      \multirow{2}{*}{Closed-source} 
        & GPT-4o~\cite{hurst2024gpt} & 61.50 & 51.00 & 65.33 & 62.00 & 68.50 & 70.00 & 74.22 & 24.50 & 14.00 & 14.00 & 25.50 \\
        & Gemini-2.5-Flash~\cite{team2024gemini} & 63.00 & 52.00 & 62.31 & 62.00 & 74.00 & 70.00 & 73.44 & 36.00 & 20.00 & 24.50 & 25.00 \\
      \midrule
      \multirow{5}{*}{Open-source} 
        & Qwen2.5-VL-7B~\cite{bai2025qwen2} & 51.50 & 45.00 & 58.00 & 54.00 & 61.00 & 61.97 & 58.14 & 12.50 & 15.00 & 18.50 & 17.00 \\
        & Qwen3-VL-8B~\cite{Qwen3-VL} & 71.50 & 48.00 & 62.50 & 63.50 & 71.50 & 61.97 & 66.67 & 27.50 & 33.00 & 21.50 & 27.50 \\
        & SpatialLadder-3B~\cite{li2025spatialladder} & 45.50 & 50.50 & 52.50 & 47.00 & 52.00 & 52.11 & 41.09 & 8.00 & 16.00 & 17.00 & 12.00 \\
        & SpatialReasoner-7B~\cite{ma2025spatialreasoner} & 38.50 & 51.50 & 62.50 & 58.50 & 59.50 & 64.79 & 54.26 & 16.00 & 36.00 & 16.50 & 13.50 \\
        & SpatialRGPT-VILA-1.5-8B~\cite{cheng2024spatialrgpt} & 50.50 & 48.00 & \cellcolor{second}{78.39} & \cellcolor{second}{69.00} & \cellcolor{second}{75.50} & 61.43 & 54.69 & 34.00 & 39.50 & 36.00 & 62.00 \\
      \midrule
      \multirow{2}{*}{Ours} 
        & N3D-VLM-3B & \cellcolor{best}{97.50} & \cellcolor{best}{95.50} & 45.50 & 66.50 & 62.50 & \cellcolor{best}{92.96} & \cellcolor{second}{93.80} & \cellcolor{second}{75.00} & \cellcolor{second}{96.00} & \cellcolor{best}{93.00} & \cellcolor{second}{83.50} \\
        & N3D-VLM-7B & \cellcolor{second}{95.00} & \cellcolor{second}{93.00} & \cellcolor{best}{90.00} & \cellcolor{best}{85.50} & \cellcolor{best}{80.00} & \cellcolor{best}{92.96} & \cellcolor{best}{95.35} & \cellcolor{best}{82.50} & \cellcolor{best}{96.50} & \cellcolor{best}{93.00} & \cellcolor{best}{87.00} \\
      \bottomrule
    \end{tabular}
  }
  % \vspace{-4pt}
  \label{tab:supple_n3d}
\end{table*}

\begin{table*}[!tp]
  \centering
  \caption{\textbf{Detailed quantitative comparison on spatial reasoning benchmark SpatialRGPT-Bench.} \cellcolor{best}{Dark green} indicates the highest accuracy, while \cellcolor{second}{light green} denotes the second-highest accuracy.}
  \vspace{-6pt}
  \resizebox{\linewidth}{!}{
    \begin{tabular}{llcccccccccccc}
      \toprule
      \multirow{3}{*}{Category} & \multirow{3}{*}{Method} & \multicolumn{12}{c}{SpatialRGPT-Bench} \\
      \cmidrule(lr){3-14}
      % Category & Method & Open-ended & Numerical & Open-ended & Numerical & Multi-choice \\
        &  & \multicolumn{6}{c}{Open-ended} & \multicolumn{6}{c}{Numerical} \\
        \cmidrule(lr){3-8} \cmidrule(lr){9-14}
        &  & below/above & left/right & big/small & tall/short & wide/thin & behind/front & distance & hor. dis. & ver. dis. & width & height & direction \\
      \midrule
      \multirow{2}{*}{Closed-source} 
        & GPT-4o~\cite{hurst2024gpt} & 88.33 & 78.10 & 82.08 & 74.11 & 68.27 & 65.45 & 30.30 & 42.31 & 53.12 & 52.24 & \cellcolor{second}71.70 & 65.42 \\
        & Gemini-2.5-Flash~\cite{team2024gemini} & 85.83 & 92.23 & 88.68 & 81.25 & 79.81 & 67.27 & 12.06 & 12.93 & 18.18 & \cellcolor{best}64.55 & \cellcolor{best}77.24 & 72.64 \\
      \midrule
      \multirow{5}{*}{Open-source} 
        & Qwen2.5-VL-7B~\cite{bai2025qwen2} & 76.67 & 89.52 & 71.70 & 71.43 & 73.08 & 64.55 & 21.15 & 28.74 & 21.62 & 35.29 & 49.50 & 57.94 \\
        & Qwen3-VL-8B~\cite{Qwen3-VL} & 93.33 & 96.19 & \cellcolor{second}88.68 & 84.82 & 88.46 & 83.64 & 29.73 & 26.23 & 29.25 & 45.80 & 55.64 & 59.05 \\
        & SpatialLadder-3B~\cite{li2025spatialladder} & 58.33 & 56.19 & 55.66 & 54.46 & 55.77 & 54.55 & 24.32 & 21.31 & 35.85 & 21.80 & 33.08 & 23.71 \\
        & SpatialReasoner-7B~\cite{ma2025spatialreasoner} & 68.33 & 48.57 & 68.87 & 66.07 & 76.92 & 50.0 & 37.16 & 34.43 & 17.92 & 34.62 & 35.34 & 42.53 \\
        & SpatialRGPT-VILA-1.5-8B~\cite{cheng2024spatialrgpt} & \cellcolor{second}99.17 & \cellcolor{best}100.0 & 84.90 & \cellcolor{second}89.28 & \cellcolor{second}91.34 & 90.90 & 45.9 & 68.0 & 56.6 & 48.9 & 61.7 & \cellcolor{second}95.3 \\
      \midrule
      \multirow{2}{*}{Ours} 
        & N3D-VLM-3B & 97.50 & \cellcolor{best}100.0 & 75.47 & 73.21 & 39.42 & \cellcolor{second}94.55 & \cellcolor{second}72.30 & \cellcolor{second}78.69 & \cellcolor{second}83.96 & 48.12 & 69.70 & 93.46 \\
        & N3D-VLM-7B & \cellcolor{best}100.0 & 99.05 & \cellcolor{best}94.34 & \cellcolor{best}92.86 & \cellcolor{best}92.31 & \cellcolor{best}95.45 & \cellcolor{best}81.08 & \cellcolor{best}85.25 & \cellcolor{best}88.68 & \cellcolor{second}54.17 & 64.29 & \cellcolor{best}96.26 \\
      \bottomrule
    \end{tabular}
  }
  % \vspace{-4pt}
  \label{tab:supple_spatialrgptbench}
\end{table*}

\subsection{Failure Cases}
\begin{figure*}[!tp]
    \centering
    \includegraphics[width=1.0\linewidth]{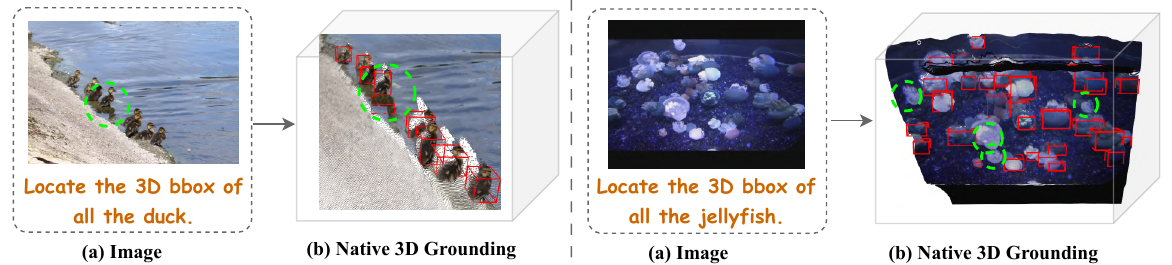}
    \vspace{-10pt}
    \caption{\textbf{Failure cases of our native 3D grounding.} N3D-VLM misidentifies a specular reflection and misses several objects, suggesting room for improvement in handling reflections and dense object scenes. }
    \label{fig:supple_fail}
    % \vspace{-10pt}
    % \vspace{-18pt}
\end{figure*}

% P0.5: figure, grouding results that fails
We present two failure cases of 3D grounding in Fig.~\ref{fig:supple_fail}. In the first example, a duck's reflection on the water surface (highlighted by a green circle) is mistakenly identified as a real object, suggesting that our model could benefit from a better understanding of specular reflections. In the second example, although N3D-VLM successfully detects 30 jellyfish with accurate 3D bounding boxes, several jellyfish are still missed, as indicated in the green circle. 
These cases highlight that our method, while effective, still has room for improvement. Enhancing the 3D grounding capabilities may further boost overall performance in 3D spatial understanding tasks. 

\section{Video Demo}
\label{sec:supple_video} 
As a supplement to Sec.~\ref{sec:supple_exp}, we have included a video demo to showcase the qualitative results in a video format.

\section{N3D-Bench Details}
\label{sec:supple_n3d}

\subsection{Distribution Summary} 
\begin{figure*}[!tp]
    \centering
    \includegraphics[width=1.0\linewidth]{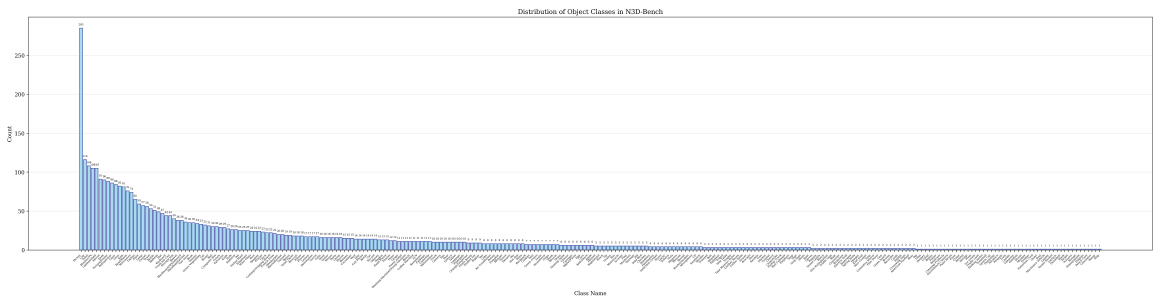}
    \vspace{-10pt}
    \caption{\textbf{Distribution of object classes in N3D-Bench.} }
    \label{fig:supple_distribution_objclass}
    \vspace{-10pt}
    % \vspace{-18pt}
\end{figure*}

\begin{figure}[t!]
    \centering
    \includegraphics[width=1.0\linewidth]{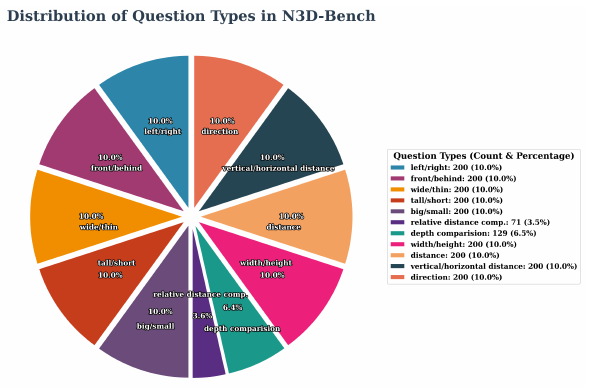}
    \vspace{-10pt}
    \caption{\textbf{Distribution of question types in N3D-Bench.} }
    \label{fig:supple_distribution_question}
    \vspace{-10pt}
    % \vspace{-18pt}
\end{figure}

% Fig.~\ref{fig:supple_distribution_question}
% \subsection{Object Class Summary} 
% Fig.~\ref{fig:supple_distribution_objclass}
In Table 1 of the main paper, we compare our N3D-Bench with SpatialRGPT-Bench, showing that N3D-Bench covers a broader range of object categories and question types. 
As a supplement to Table 1, we further present the distribution of question types and object categories in N3D-Bench. Specifically, Fig.~\ref{fig:supple_distribution_question} shows the distribution of question types. The dataset contains 2,000 questions grouped into 11 major categories, reflecting both high diversity and balanced coverage. For example, the questions include reasoning about \emph{relative distance comparison} and \emph{depth comparison} among three or more objects. For relative spatial relations such as \emph{left/right}, we also incorporate variations in viewpoint, including phrasing like “from the opposite direction of the camera.”
Fig.~\ref{fig:supple_distribution_objclass} illustrates the distribution of object categories involved in the questions. N3D-Bench includes 264 commonly encountered indoor and outdoor object classes, derived from the Objects365~\cite{shao2019objects365} dataset.

\clearpage
% \newpage
